\documentclass[runningheads]{llncs}

\usepackage{graphicx}
\usepackage{comment}
\usepackage{amsmath,amssymb}
\usepackage{color}
\usepackage{url}
\usepackage{hyperref}
\usepackage{booktabs}
\usepackage{enumitem}

\newif\ifreview
\reviewfalse

\ifreview
	\usepackage{lineno}

	\linenumbers
\fi

\begin{document}


\def\SubNumber{106}

\def\GCPRTrack{Main Track}

\title{Combined Image Data Augmentations diminish the benefits of Adaptive Label Smoothing}

\ifreview
\else
	\titlerunning{Aggressive Soft Augmentation}

	\author{Georg Siedel \inst{1,2}\orcidID{0009-0004-6190-2726} \and
	Ekagra Gupta \inst{2} \and
	Weijia Shao \inst{1} \and
	Silvia Vock \inst{1}\and
	Andrey Morozov \inst{2}}
	
	\authorrunning{Siedel et al.}
	
	\institute{Federal Institute for Occupational Safety and Health (BAuA), Dresden, Germany \email{\{siedel.georg,shao.weijia,vock.silvia\}@baua.bund.de} \and University of Stuttgart, Germany}

\fi

\maketitle              

\begin{abstract}

Soft augmentation regularizes the supervised learning process of image classifiers by reducing label confidence of a training sample based on the magnitude of random-crop augmentation applied to it. This paper extends this adaptive label smoothing framework to other types of aggressive augmentations beyond random-crop. Specifically, we demonstrate the effectiveness of the method for random erasing and noise injection data augmentation. Adaptive label smoothing permits stronger regularization via higher‐intensity Random Erasing. However, its benefits vanish when applied with a diverse range of image transformations as in the state-of-the-art TrivialAugment method, and excessive label smoothing harms robustness to common corruptions. Our findings suggest that adaptive label smoothing should only be applied when the training data distribution is dominated by a limited, homogeneous set of image transformation types.

\end{abstract}

\section{Introduction}

\begin{figure}[t]
    \centering
    \includegraphics[width=\linewidth,trim={10 118 10 8},clip]{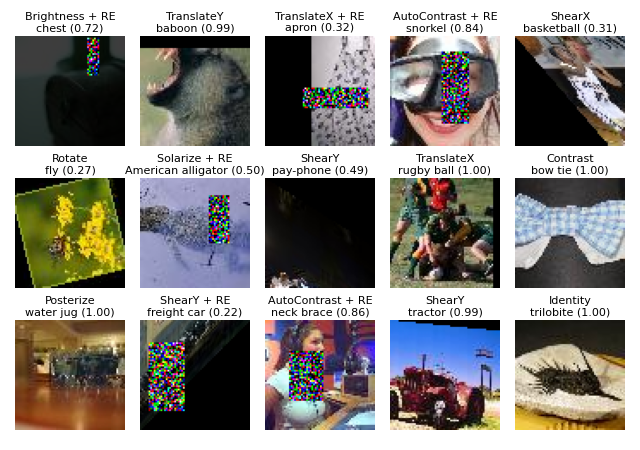}
    \caption{TinyImageNet images transformed with soft TrivialAugment and soft Random Erasing (RE) data augmentation. The titles display the TrivialAugment transformation type and whether RE is applied, as well as the label and its softened confidence. The confidence is calculated as a function of the augmentation severity for every transformation type individually. Here, the functions are derived from a proxy models average accuracy on that transformation type and severity, as can be seen from Figure \ref{fig:mapping_all}.}
    \label{fig:example_images}
\end{figure}

Vision models have long surpassed human accuracy on tasks such as image classification \cite{krizhevsky2012imagenet}. A key success factor to their ability to learn general and robust representations of data is data augmentation, which enriches training data diversity by applying controlled transformations to images or labels \cite{Shorten2019}. Among image data augmentations, very aggressive transformations have proved effective in methods such as TrivialAugment \cite{Muller2021}. However, excessively transforming images can lead to information loss in the image (see Figure \ref{fig:example_images}). In this case, a model will learn confident labels from uncertain information, which could lead to its miscalibration \cite{liu2023soft}. Soft augmentation \cite{liu2023soft} addresses this issue by combining image augmentation with adaptive label smoothing. The method scales down label confidences the higher the magnitude of image transformations become. Specifically applied to random cropping augmentations, this enabled more aggressive crops during training without overfitting or model miscalibration. 

In this work, we extend adaptive label smoothing to well-established, aggressive image augmentation strategies like TrivialAugment and Random Erasing to further enhance their efficacy, as demonstrated by the image examples in Figure \ref{fig:example_images}.
Our contributions can be summarized as follows:\footnote{Code available at \href{https://github.com/Georgsiedel/soft_label_random_augmentation}{https://github.com/georgsiedel/soft\_label\_random\_augmentation}}

\begin{itemize}[wide]
    \item We integrate adaptive label smoothing with TrivialAugment. We model magnitude-confidence mapping functions for its various transformation types from human vision studies, proxy networks, and image similarity metrics.
    \item Through image classification experiments, we show that adaptive label smoothing enables more aggressive parameter settings for Random Erasing and can improve the performance of noise injection data augmentation.
    \item We demonstrate that the benefit of adaptive label smoothing vanishes when applied across a heterogeneous set of augmentations as in TrivialAugment, which constrains its practical applicability.
\end{itemize}

\section{Related Work}

\subsection{Image Augmentation}
Early geometric image augmentations such as random flips and random crops (RC) \cite{krizhevsky2012imagenet,Zagoruyko2016} are now standard in training pipelines. Occlusion methods such as Cutout \cite{devries2017improved} and Random Erasing (RE) \cite{Zhong2020randomerasing} eliminate image regions to force reliance on global context. Noise injections, ranging from Gaussian Noise \cite{Rusak2020} over general p-norm noise \cite{Siedel2024} to Patch Gaussian noise \cite{Lopes2019}, randomizes pixel‐level statistics and induces robustness against high-frequency corruptions. 

AutoAugment and RandAugment \cite{cubuk2019autoaugment,cubuk2020randaugment} were pioneering approaches for more advanced combination policies over a set of fourteen transformations types. The latest TrivialAugment (TA) \cite{Muller2021} simplified this approach using the same transformation types: it randomly selects one transform and a more aggressive magnitude compare to its predecessors. TA is tuning-free and matches or outperforms AutoAugment and RandAugment.

These input-only augmentation schemes are state-of-the-art in classification accuracy and robustness to common corruptions \cite{Vryniotis2021,Erichson2024}. They presume that important image content, in particular the class labels, remain invariant under all transformations, but aggressive distortions as in TA can violate this, as displayed in Figure \ref{fig:example_images}, yielding overconfident or miscalibrated predictions \cite{liu2023soft}. This shortcoming motivates a complementary line of research: augmenting the labels themselves to better represent the actual transformed image data.

\subsection{Label Augmentation}
Label augmentation alters ground-truth targets in the training data. Early methods include randomly flipping labels \cite{xie2016disturblabel}, and Label Smoothing, which redistributes a fraction $\alpha$ of the probability mass uniformly across non‐target classes \cite{szegedy2016rethinking}. While effective at regularizing, these approaches apply to every sample in an identical way.

Adaptive label augmentation tailors the label smoothing to the image data. Online Label Smoothing derives targets from the model’s own predicted class confidences at each iteration \cite{zhang2021delving}. Label Refinery employs a teacher network to generate softer labels that better reflect each image’s content \cite{bagherinezhad2018label}. Meta‐learning–based schemes estimate optimal labels per sample via an auxiliary objective \cite{vyas2020learning}.

Other methods jointly augment inputs and labels. Mixup and Cutmix interpolate paired images and their labels, enforcing linear behavior between classes and improving accuracy, robustness and model calibration \cite{Zhang2018mixup,yun2019cutmix}. However, neither deploy image transformations to add additional diversity to the training process.

\subsection{Soft Augmentation}
Co‐designing input and label perturbations lets a model learn diverse distortions while calibrating its confidence to their severity. In \textit{soft augmentation} \cite{liu2023soft}, each image-label-pair $(x_i, y_i)$ is transformed as

\begin{equation}\label{eq:soft_augmentation}
    \underbrace{\Big(x_i, y_i\Big)}_{\text{Training data tuple}} \mapsto \quad \underbrace{\Big( t_{\phi \sim S}(x_i)}_{\text{Image augmentation}} \:,\:\underbrace{g_\alpha(\phi)(y_i)\Big)}_{\text{Adaptive label smoothing}}
\end{equation}

where $\phi \sim S$ denotes the transformation type and magnitude sampled from a set $S$, which informs the factor $\alpha$ of the label smoothing function $g_\alpha(y_i)$. Intuitively, stronger augmentations yield lower target confidence. An additional reweighting \cite{ren2018learning} of the loss with the reduced confidence further amplifies the smoothing regularization. 
Liu et al. \cite{liu2023soft} show that this adaptive scheme enables more aggressive image augmentation without overfitting, improving both accuracy and calibration. Their work, however, is limited to random cropping (RC).

\subsection{Human Vision Studies}
The soft RC augmentation is inspired by human visual perception, long the gold standard for vision models \cite{serre2007feedforward,zhu2019robustness}. Liu et al. \cite{liu2023soft} map the reduction in label confidence to the magnitude of RC on an image by adapting data from a human vision study (HVS). The study measures classification accuracy on occluded image and finds that human performance remains high under mild to moderate occlusion but degrades nonlinearly toward chance level as occlusion intensifies \cite{tang2018recurrent}.

Comparable HVS for other image transformations are available for noise and blur \cite{dodge2019human}, contrast variation \cite{avidan2002contrast}, and rotation \cite{hollard1982rotational,Kail1985development}. Note, however, that many psychophysical experiments report response time rather than classification accuracy \cite{gardner2005contrast,mazade2022cortical}, making them unsuitable proxies for mapping label confidence.

\section{Adaptive Label Smoothing for Aggressive Image Data Augmentation}

In this paper, we extend the soft augmentation from Equation \ref{eq:soft_augmentation} from Random Cropping (RC) to other, aggressive image augmentation schemes in order to find out if the method transfers. However, for every new transformation type in an image augmentation scheme, we need to model functions that map transformation magnitude to smoothed label confidence in a reasonable way. To this end, several approaches can be used to estimate how much information is lost in the image due to transformation of this type with certain magnitude.

\subsection{Mapping label confidence to image transformation magnitude}

\begin{figure}[t]
    \centering
    \includegraphics[width=0.85\linewidth,trim={7 16 7 7},clip]{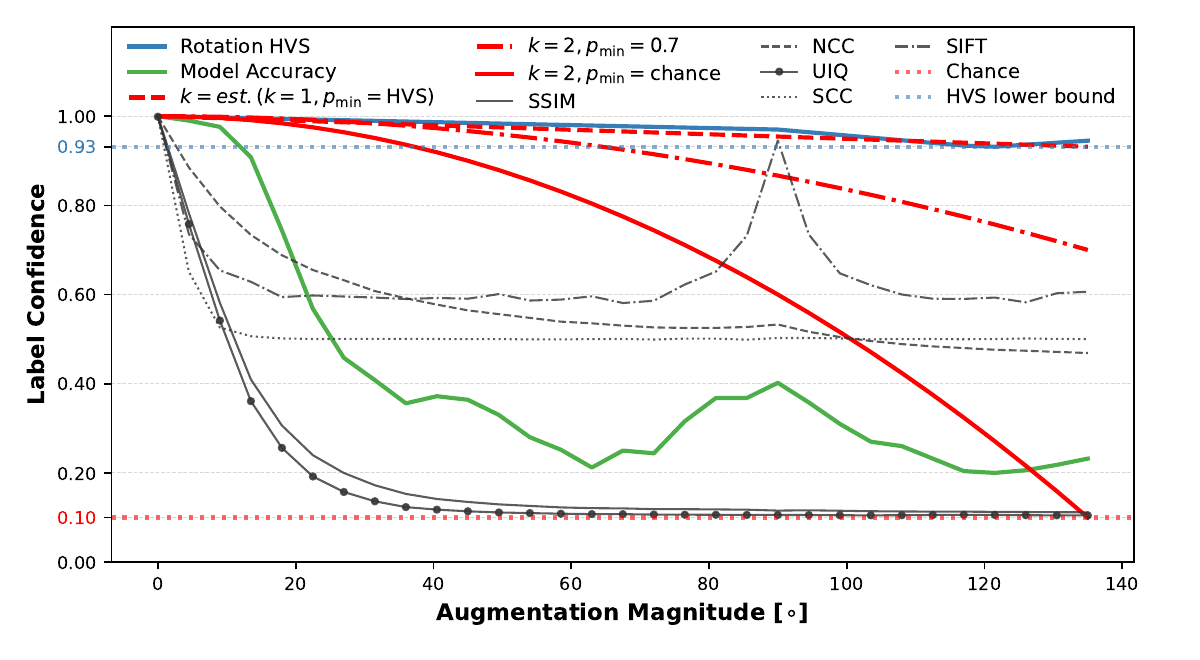}
    \caption{This diagram displays the functions defining how an images label is adjusted based on the magnitude of the image transformation. For the "Rotate" transformation in TrivialAugment, the functions map label confidence to image rotation in degrees. The functions are based on human vision studies (blue), on a proxy models outputs (green), on custom polynomial functions (red) and on image similarity metrics (black). Here, functions differ significantly between all approaches, emphasizing the potential differences in modeling approaches for adaptive label smoothing. The polynomial function $k=est.$ mimics HVS data if that is available like in for image rotations, otherwise it mimics model accuracy.}
    \label{fig:mapping_rotation}
\end{figure}

This section describes such approaches used in this study. Figure \ref{fig:mapping_rotation} shows the mapping functions for those approaches exemplarily for the rotation transformation.

\paragraph{HVS}
Wherever available, we adopt HVS data to set label confidence to average human classification accuracy under a given distortion level, as in \cite{liu2023soft}. If the HVS covers only parts of the magnitude range, we linearly interpolate between measured points. Transformations lacking any HVS accuracy data get no smoothing under what we denote the $\mathrm{HVS}$ scheme in the following.

\paragraph{Model‐accuracy}
Inspired by \cite{bagherinezhad2018label,vyas2020learning,zhang2021delving}, we derive what we denote a model-accuracy mapping from a classifier’s accuracy on CIFAR-10 images. The model has been pretrained with Mixup and noise injections, but with no transformations contained in TA or RE. We measure the average model accuracy on training images subjected to the transformations to be mapped. This setup corresponds to a model that knows all images, but is confronted with unknown transformations. Starting from 100 \%, accuracy falls depending on transformation magnitude, which we use as a fixed label confidence mapping. 

\paragraph{Image comparison metrics} Alternatively, we explored image comparison metrics to quantify distortion severity. We applied Structural Similarity Index Measure (SSIM) \cite{wang2004image}, (Fast) Normalized Cross Correlation (NCC) \cite{yoo2009fast}, Spatial Correlation Coefficient (SCC) \cite{brown2002invariant}, Universal Image Quality (UIQ) \cite{wang2002universal} and Scale-Invariant Feature Transform (SIFT) \cite{lowe2004distinctive} to 500 image pairs at each transformation magnitude. All metrics except SIFT yield scores in $[-1,1]$, which we naively rescaled to $[chance,1]$. For SIFT, which is used only on geometric transforms, we measured the fraction of tracked keypoints retained through a transformation. As shown in Figures \ref{fig:mapping_rotation} and \ref{fig:mapping_all} for the transformation types contained in TA, NCC and SIFT align with HVS and model‐accuracy data on geometric transformations, whereas SSIM, UIQ, and SCC better reflect color transformations. However, no single metric consistently behaves intuitively and comparably to HVS or model accuracy across the transformation types. Since it appeared far-fetched to assemble a mapping based on a manually crafted mixture of metrics, we omitted the image comparison approach from our training experiments.

\paragraph{Polynomial estimate} As a parametric fallback, we also tested a simple polynomial mapping:
\[
\alpha(\phi) = \phi^{k}\,(1 - p_{\min}),
\]
where $\phi \in [0,1]$ is the transformation magnitude (e.g. occluded image ratio, rotation angle with maximum rotation normalized to 1), $p_{\min}$ is the minimum confidence and $k$ controls the functions curvature.  

We evaluate four polynomial variants:
\begin{enumerate}
  \item \textbf{$k=2,\ p_{\min} = chance$:} Baseline as in \cite{liu2023soft}. $chance=1/classes$, so that $p_{\min}$ floors label smoothing at random-guessing level.  
  \item \textbf{$k=2,\ p_{\min} = 0.7$:} Yields an average label smoothing factor of 0.1 as is standard \cite{szegedy2016rethinking}, given uniformly distributed $\phi$.  
  \item \textbf{$k=est.$:} $k$ and $p_{\min}$ individually approximate the HVS curve (or the model‐accuracy mapping when HVS is unavailable).
  \item \textbf{$k=2,\ p_{\min} = 0.3$:} Only for noise, based on HVS in \cite{dodge2019human}.
\end{enumerate}

\subsection{Soft TrivialAugment}

Next, we select data augmentation candidates to which adaptive label smoothing should be extended. TA is a natural candidate for softening due to its remarkable performance and the aggressive image transformations it employs. This method randomly selects from 14 transformation types and 31 discrete magnitudes, uniformly distributed over predefined intervals specific to each type. 
Table \ref{tab:TA_transforms} summarizes all transformation types in TA, their respective data sources for the HVS mapping (where available), and the data used for the $k=est.$ mapping.

\begin{figure}[ht!]
    \centering
    \includegraphics[width=\linewidth,trim={7 6.5 7 13.5},clip]{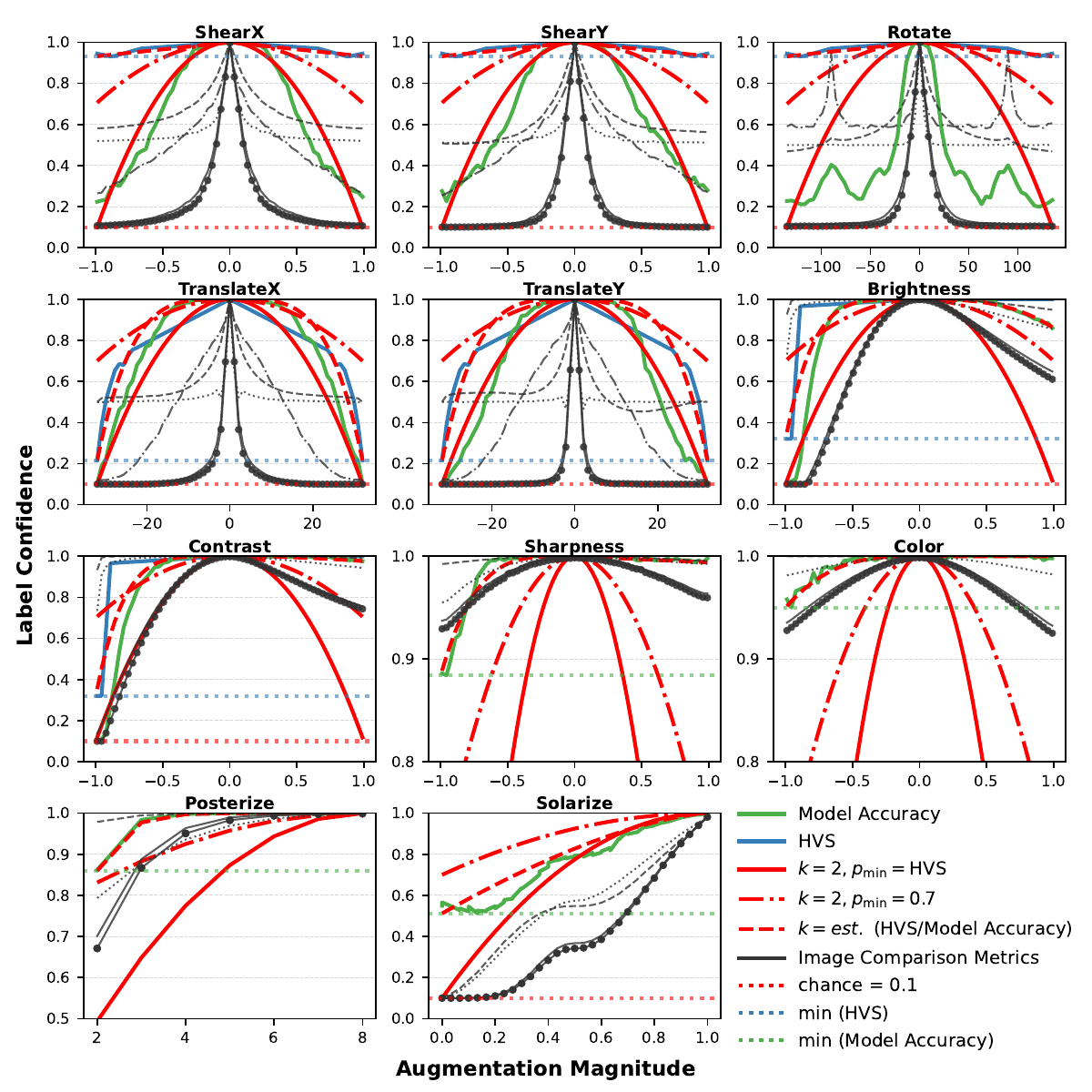}
    \caption{A summary of the functions defining how an images label is adjusted based on the magnitude of the image transformation, for all transformation types used in TrivialAugment (find a closer view for one transformation type in Figure \ref{fig:mapping_rotation}. The functions are based on human vision studies (HVS, blue), on a proxy models outputs (green), on custom polynomial functions (red) and on image similarity metrics (black). The polynomial function $k=est.$ mimics HVS data if available,  model accuracy otherwise.}
    \label{fig:mapping_all}
\end{figure}

Figure \ref{fig:mapping_all} summarizes all available mapping functions as described in detail earlier, for those 11 transformations in TA that have variable magnitudes.

\setlength{\tabcolsep}{20pt}
\begin{table}[t]
\caption{All TrivialAugment transformations and the human vision study data used for mapping the label confidence to the transformation magnitude for this transformations. We transfer human vision study data from Rotation to Shear and from Contrast to Brightness, because we found the proxy models behaviour and several image similarity metrics to behave similarly on these transformations, in the same manner as \cite{liu2023soft} use Occlusion studies for Translation transformations. Four transformations have no human vision study data available that fits this transformation. The last column shows whether the $k=est.$ mapping mimics the human vision study data or model accuracy. The last three TrivialAugment transformations have no mapping for adaptive label smoothing at all, as they are no transformation (Identity) or non-variable in magnitude (Equalize and AutoContrast).}\label{tab:TA_transforms}
\resizebox{\textwidth}{!}{%
\begin{tabular}{@{}lcc}
\hline
TrivialAugment Transformation &  Human Vision Study mapping & $k=est.$ mapping\\
\hline
Rotate & Rotation \cite{hollard1982rotational,Kail1985development} & HVS\\
ShearX & Rotation \cite{hollard1982rotational,Kail1985development} & HVS\\
ShearY & Rotation \cite{hollard1982rotational,Kail1985development} & HVS\\
TranslateX & Occlusion \cite{tang2018recurrent} & HVS\\
TranslateY & Occlusion \cite{tang2018recurrent} & HVS\\
Brightness & Contrast \cite{avidan2002contrast} & HVS\\
Contrast & Contrast \cite{avidan2002contrast} & HVS \\
Sharpness & - & Model Accuracy\\
Color & - & Model Accuracy\\
Posterize & - & Model Accuracy\\
Solarize & - & Model Accuracy\\
Equalize, AutoContrast, Identity & \multicolumn{2}{c}{-}\\
\hline
\end{tabular}}
\end{table}

\subsection{Soft Random Erasing}

We extend adaptive label smoothing to RE \cite{Zhong2020randomerasing}, an occlusion-based augmentation method, hence it may benefit from adaptive label smoothing similarly to RC. Following the methodology in Liu et al. \cite{liu2023soft} and occlusion-based HVS data in \cite{tang2018recurrent}, we adopt the polynomial mapping function $k=2,\ p_{\min} = chance$, where the transformation magnitude is the occluded area ratio of the image.

\subsection{Soft Noise}

Last, we apply adaptive label smoothing to two noise injection variants: Gaussian noise is drawn from $N(0,0.1)$ and applied across the entire image. Patch Gaussian noise is drawn from $N(0,1.0)$ and applied to a fixed square region (25 pixels, 50 for TinyImageNet) centered randomly on the image as in \cite{Lopes2019}.

In both cases, noise is scaled by a random factor drawn uniformly from $[0,1.0]$ per image. The transformation magnitude is defined as the product of the Gaussian standard deviation, the scaling factor, and the relative area affected by the noise. Label confidence is then computed using the polynomial mapping function $k=2,\ p_{\min} = 0.3$, as estimated from the HVS study in \cite{dodge2019human}.

\section{Experiments}\label{sec:experiments}

We test our approach on the image classification datasets CIFAR-10 (C10), CIFAR-100 (C100) \cite{Krizhevsky2009}, and TinyImageNet (TIN) \cite{Le2015}, primarily using a WideResNet-28-4 backbone \cite{Zagoruyko2016} or a ResNeXt-29-32x4d \cite{Xie2017resnext} for validation. Training hyperparameters are detailed in Appendix A.

Beyond test accuracy, we evaluate the corruption robustness of models. The corruption robustness of a classifier $f : X \to Y$ according to \cite{Hendrycks2019a} is: 

\begin{equation}\label{eq:corr_rob}
\mathbb{E}_{c \sim C} \left[ P_{(x, y) \sim D} \big(f(c(x)) = y \big) \right]
\end{equation}

where $x,y$ are data samples from the data distribution $D$. $C$ is a set of corruptions which perturb $x$. Here, $C$ is the corruption robustness benchmark from \cite{Hendrycks2019a} that contains 19 real-world corruptions in 5 severities. Robustness denotes the average accuracy across all corruptions of this benchmark dataset on all test images.

\subsection{Analysis of different mappings approaches and transformation types in soft TrivialAugment}

\setlength{\tabcolsep}{5pt}

\begin{table}[t]
  \centering
  \caption{Mean accuracy and robustness with standard deviations over 5 runs for different soft TrivialAugment mapping approaches compared to hard label TrivialAugment and TrivialAugment with standard label smoothing (with 0.1 label weight redistributed). (w) denotes a reweighted variant of the most promising mapping $k=2, p_{min}=0.7$, while the second block validates this mappings slight advantage on CIFAR datasets on the ResNeXt model architecture. Note that no improvement over the baseline is larger than the combined standard deviations, indicating little statistical significance.}
  \label{tab:TA-mapping-comparison}
  \resizebox{\textwidth}{!}{%
  \begin{tabular}{@{}l  cc  cc  cc@{}}
    \toprule
    & \multicolumn{2}{c}{CIFAR-10}
    & \multicolumn{2}{c}{CIFAR-100}
    & \multicolumn{2}{c}{TinyImageNet} \\
    \cmidrule(lr){2-3}\cmidrule(lr){4-5}\cmidrule(lr){6-7}

    Mapping 
      & Acc.      & Rob.
      & Acc.      & Rob.
      & Acc.      & Rob. \\
    \midrule

    \addlinespace[3pt]
    \multicolumn{7}{@{}l}{\small\bfseries WRN-28-4} \\[3pt]

    Hard Labels
      & 96.62 {\scriptsize$\pm$0.16} 
      & 87.12 {\scriptsize$\pm$0.43}
      & 80.20 {\scriptsize$\pm$0.25}
      & 62.55 {\scriptsize$\pm$0.64}
      & 65.44 {\scriptsize$\pm$0.28}
      & 31.02 {\scriptsize$\pm$0.94} \\

    Label Smoothing
      & 96.66 {\scriptsize$\pm$0.07}
      & 86.91 {\scriptsize$\pm$0.56}
      & 80.30 {\scriptsize$\pm$0.14}
      & \textbf{63.31} {\scriptsize$\pm$0.46}
      & \textbf{65.52} {\scriptsize$\pm$0.18}
      & \textbf{32.56} {\scriptsize$\pm$0.82} \\ [3pt]

    Model Accuracy
      & 96.50 {\scriptsize$\pm$0.17}
      & \textbf{87.48} {\scriptsize$\pm$0.35}
      & 79.70 {\scriptsize$\pm$0.39}
      & 62.73 {\scriptsize$\pm$0.61}
      & 64.02 {\scriptsize$\pm$0.48}
      & 30.64 {\scriptsize$\pm$0.86} \\

    $HVS$
      & 96.55 {\scriptsize$\pm$0.12}
      & 87.22 {\scriptsize$\pm$0.31}
      & 80.16 {\scriptsize$\pm$0.31}
      & 62.74 {\scriptsize$\pm$0.51}
      & 64.90 {\scriptsize$\pm$0.69}
      & 30.20 {\scriptsize$\pm$1.06} \\

    $k=est.$
      & 96.58 {\scriptsize$\pm$0.20}
      & 87.20 {\scriptsize$\pm$0.44}
      & 80.26 {\scriptsize$\pm$0.23}
      & 62.38 {\scriptsize$\pm$0.43}
      & 64.88 {\scriptsize$\pm$0.20}
      & 30.85 {\scriptsize$\pm$1.80} \\

    $k=2, p_{min}=chance$
      & 96.42 {\scriptsize$\pm$0.15}
      & 85.37 {\scriptsize$\pm$0.89}
      & 79.66 {\scriptsize$\pm$0.17}
      & 60.01 {\scriptsize$\pm$0.65}
      & 63.95 {\scriptsize$\pm$0.53}
      & 26.42 {\scriptsize$\pm$2.05} \\

    $k=2, p_{min}=0.7$
      & \textbf{96.78} {\scriptsize$\pm$0.08}
      & 86.93 {\scriptsize$\pm$0.32}
      & \textbf{80.37} {\scriptsize$\pm$0.12}
      & 62.35 {\scriptsize$\pm$0.36}
      & 65.08 {\scriptsize$\pm$0.37}
      & 30.38 {\scriptsize$\pm$1.99} \\

    $k=2, p_{min}=0.7$ (w)
      & 96.60 {\scriptsize$\pm$0.15}
      & 86.83 {\scriptsize$\pm$0.18}
      & 80.08 {\scriptsize$\pm$0.32}
      & 61.60 {\scriptsize$\pm$0.75}
      & 65.25 {\scriptsize$\pm$0.39}
      & 27.95 {\scriptsize$\pm$1.84} \\

    \addlinespace[8pt]
    \multicolumn{7}{@{}l}{\small\bfseries ResNeXt-29-32x4d} \\[3pt]
    Hard Labels
      & 96.43{\scriptsize$\pm$0.06}
      & 86.50{\scriptsize$\pm$0.39}
      & 80.12{\scriptsize$\pm$0.19}
      & 61.44{\scriptsize$\pm$0.86}
      & \textbf{67.13}{\scriptsize$\pm$0.18}
      & \textbf{32.05}{\scriptsize$\pm$0.72}\\ [2pt]

    $k=2, p_{min}=0.7$
      & \textbf{96.61}{\scriptsize$\pm$0.15}
      & \textbf{86.78}{\scriptsize$\pm$0.54}
      & \textbf{80.64}{\scriptsize$\pm$0.52}
      & \textbf{61.76}{\scriptsize$\pm$0.50}
      & 67.06{\scriptsize$\pm$0.65}
      & 29.83{\scriptsize$\pm$0.83}\\

    \bottomrule
  \end{tabular}}
\end{table}

\begin{figure}[!ht]
    \centering
    \includegraphics[width=\linewidth,trim={8 13 8 10.3},clip]{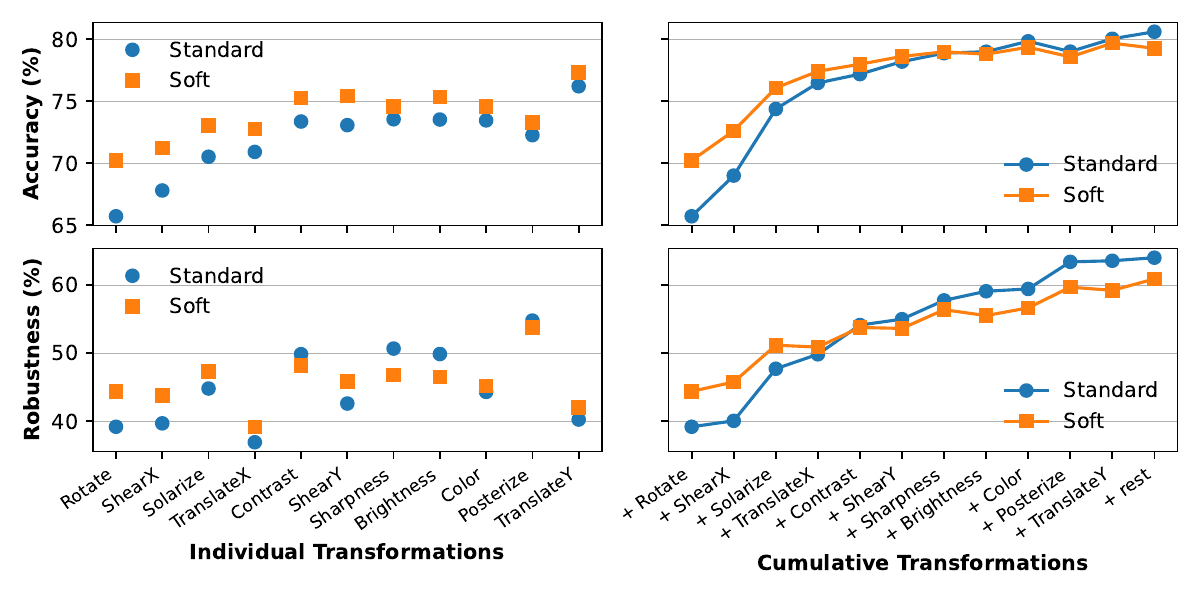}
    \caption{Standard TrivialAugment and TrivialAugment with adaptive label smoothing (soft) are compared for individual transformation types from its set (left) and when incrementally adding transformations (right), evaluating accuracy (top) and robustness (bottom). The transformations are ordered along the x-axis by decreasing incremental benefit of adaptive label smoothing. The $k=2, p\geq chance$ mapping is used.}
    \label{fig:TA-individual}
\end{figure}

Table \ref{tab:TA-mapping-comparison} compares different soft TA mapping functions to TA and TA with fixed label smoothing. The conservative $k=2, p\geq 0.7$ mapping yields a small gain over TA and fixed label smoothing on CIFAR, but not on TIN. All aggressive mappings ($k=2, p\geq chance$, model accuracy, weighted) degrade performance. Any observed improvements lie within the standard deviations across 5 runs, indicating little statistical significance.

To diagnose the limited overall benefit of soft TA, we isolate each transformation type in TA and retrain with adaptive label smoothing. Figure \ref{fig:TA-individual} shows that adaptive label smoothing does boosts accuracy on individual augmentations. However, as transformations are incrementally accumulated, the gains vanish and eventually reverse, particularly deteriorating robustness to common corruptions. Even when we use all transformations, but apply adaptive label smoothing only to the 3 transformation types that benefit the most from it, the model accuracy is no better than for standard TA.

\subsection{More aggressive parametrization of soft Random Erasing}\label{sec:RE-params}
\begin{figure}[t]
    \centering
    \includegraphics[width=\linewidth,trim={8 3.5 8 9},clip]{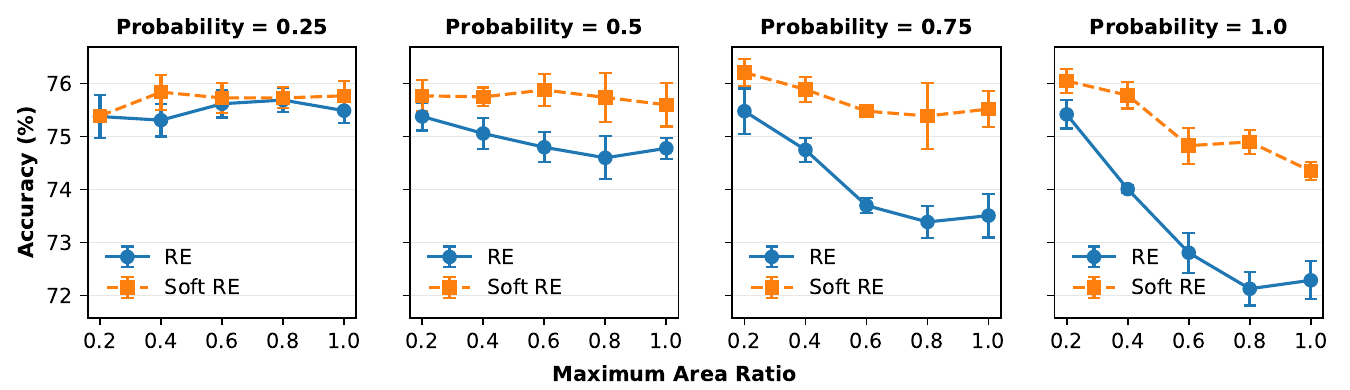}
    \caption{Parameter sweep of the two main hyperparameters of Random Erasing on CIFAR-100. Displayed are the mean and standard deviations for accuracy over 5 runs for standard and soft Random Erasing. No reweighting or other augmentations are applied.}
    \label{fig:re_sweep}
\end{figure}

Inspired by soft RC's tolerance for stronger distortions in \cite{liu2023soft}, we perform a grid sweep for standard and soft RE over its two key hyperparameters - application probability and maximum area ratio, which is the upper bound to the randomly drawn occluded image proportion. As shown in Figure \ref{fig:re_sweep}, soft RE consistently outperforms its hard‐label counterpart. At a low application probability of 0.25, the improvement is modest. However, it increases substantially as both the probability and the size of the occlusion grow, confirming the effectiveness of adaptive label smoothing in allowing for stronger image distortions. In all following experiments, we use the optimal parameters found for both methods.

\subsection{Soft Random Erasing and Noise Injections}

Following our soft TA analysis, we expect the single‐transform augmentations RE, Gaussian and Patch Gaussian to benefit from adaptive label smoothing much like RC. Table \ref{tab:soft-single-augs} confirms that models trained with soft RE or soft noise generally outperform their hard-label counterparts, both alone and on top of TA. Soft RE on TIN is a lone exception - the reason likely is that we did not parametrize RE on TIN, but reused the parameters for C100 from section \ref{sec:RE-params}, which may not transfer optimally. Noise injections show an overall tendency to improve with soft labels, though not uniformly across experiments and not by large margins. Soft RC stands out as the most effective method for improving accuracy, but it consistently hurts robustness. Patch Gaussian (in most cases soft) offers the best tradeoff between accuracy and robustness. Contrary to findings by Liu et al. \cite{liu2023soft}, reweighting does not benefit training in any experiment.

In Table \ref{tab:ablation} we show ablation studies where augmentations are incrementally applied, with and without adaptive label smoothing, on top of TA. The results show that RC and RE can be combined and, on TIN, still benefit from softening, even when applied on top of TA. For Gaussian and Patch Gaussian noise injections, softening is unfavorable, likely due to the fact that in this ablation study, noise is the third consecutive augmentation. Here, the reduced confidence is multiplied for all soft augmentations, so the label smoothing extend may become excessive.
With respect to robustness, softening is unfavorable almost across the board. Patch Gaussian offers a more favorable tradeoff between accuracy and robustness on CIFAR, but its parametrization appears less applicable  for TIN.

\setlength{\tabcolsep}{6pt}

\begin{table}[t]
  \centering
  \caption{Single‐transform augmentations vs. their soft versions, applied individually and on top of TrivialAugment. Standard deviations are reported over 5 runs - notice that not all improvements can be considered statistically significant. P.-Gaussian means Patch Gaussian, (w) highlight experiments with reweighted labels. Only soft Random Crop boosts accuracy by significant margins consistently across datasets. Soft Random Erasing and soft Patch Gaussian work well on CIFAR, but not on TinyImageNet.}
  \label{tab:soft-single-augs}
  \resizebox{\textwidth}{!}{%
  \begin{tabular}{@{}l  cc  cc  cc@{}}
    \toprule
    & \multicolumn{2}{c}{CIFAR-10}
    & \multicolumn{2}{c}{CIFAR-100}
    & \multicolumn{2}{c}{TinyImageNet} \\
    \cmidrule(lr){2-3}\cmidrule(lr){4-5}\cmidrule(lr){6-7}
    Method 
      & Acc.      & Rob.
      & Acc.      & Rob.
      & Acc.      & Rob. \\
    \midrule
    Baseline
      & 93.92 {\scriptsize$\pm$0.11}
      & 72.98 {\scriptsize$\pm$0.24}
      & 73.87 {\scriptsize$\pm$0.33}
      & 45.52 {\scriptsize$\pm$0.61}
      & 58.83 {\scriptsize$\pm$0.34}
      & 22.29 {\scriptsize$\pm$0.69} \\[4pt]

    P.-Gaussian
      & 94.32 {\scriptsize$\pm$0.08}
      & 81.37 {\scriptsize$\pm$0.56}
      & 73.71 {\scriptsize$\pm$0.28}
      & 55.33 {\scriptsize$\pm$0.40}
      & 58.34 {\scriptsize$\pm$0.33}
      & 23.17 {\scriptsize$\pm$0.63} \\

    P.-Gaussian (Soft)
      & 94.84 {\scriptsize$\pm$0.12}
      & 82.70 {\scriptsize$\pm$0.76}
      & 73.50 {\scriptsize$\pm$0.31}
      & 55.74 {\scriptsize$\pm$0.45}
      & 58.26 {\scriptsize$\pm$0.21}
      & 23.29 {\scriptsize$\pm$0.55} \\[2pt]

    Gaussian
      & 91.66 {\scriptsize$\pm$0.21}
      & 83.29 {\scriptsize$\pm$0.41}
      & 68.85 {\scriptsize$\pm$0.17}
      & \textbf{57.19} {\scriptsize$\pm$0.32}
      & 57.71 {\scriptsize$\pm$0.35}
      & 23.86 {\scriptsize$\pm$0.82} \\

    Gaussian (Soft)
      & 91.84 {\scriptsize$\pm$0.15}
      & \textbf{83.32} {\scriptsize$\pm$0.22}
      & 68.70 {\scriptsize$\pm$0.44}
      & 57.06 {\scriptsize$\pm$0.29}
      & 56.87 {\scriptsize$\pm$0.55}
      & \textbf{23.97} {\scriptsize$\pm$0.68} \\[2pt]

    RE
      & 94.86 {\scriptsize$\pm$0.13}
      & 74.19 {\scriptsize$\pm$0.29}
      & 75.69 {\scriptsize$\pm$0.25}
      & 47.77 {\scriptsize$\pm$0.48}
      & 60.52 {\scriptsize$\pm$0.29}
      & 20.83 {\scriptsize$\pm$0.73} \\

    RE (Soft)
      & 94.98 {\scriptsize$\pm$0.11}
      & 74.32 {\scriptsize$\pm$0.99}
      & 76.21 {\scriptsize$\pm$0.28}
      & 47.96 {\scriptsize$\pm$0.60}
      & 60.05 {\scriptsize$\pm$0.26}
      & 21.15 {\scriptsize$\pm$0.53} \\

    RE (Soft) (w)
      & 94.95 {\scriptsize$\pm$0.06}
      & 74.71 {\scriptsize$\pm$0.56}
      & 75.98 {\scriptsize$\pm$0.44}
      & 48.30 {\scriptsize$\pm$0.64}
      & 60.16 {\scriptsize$\pm$0.24}
      & 21.69 {\scriptsize$\pm$0.38} \\[2pt]

    RC
      & 95.09 {\scriptsize$\pm$0.11}
      & 73.40 {\scriptsize$\pm$0.39}
      & 76.58 {\scriptsize$\pm$0.21}
      & 46.10 {\scriptsize$\pm$0.55}
      & 60.62 {\scriptsize$\pm$0.18}
      & 21.95 {\scriptsize$\pm$0.81} \\

    RC (Soft)
      & \textbf{95.99} {\scriptsize$\pm$0.11}
      & 71.94 {\scriptsize$\pm$0.59}
      & \textbf{78.01} {\scriptsize$\pm$0.33}
      & 43.49 {\scriptsize$\pm$0.58}
      & \textbf{62.06} {\scriptsize$\pm$0.57}
      & 19.96 {\scriptsize$\pm$0.91} \\

    RC (Soft) (w)
      & 95.92 {\scriptsize$\pm$0.22}
      & 70.98 {\scriptsize$\pm$0.40}
      & 77.43 {\scriptsize$\pm$0.32}
      & 43.33 {\scriptsize$\pm$0.62}
      & 61.70 {\scriptsize$\pm$0.71}
      & 20.52 {\scriptsize$\pm$0.84}\\[4pt]
    
    \cline{2-7}
    \addlinespace[3.5pt]  
    
    TrivialAugment
      & 96.42 {\scriptsize$\pm$0.07}
      & 86.78 {\scriptsize$\pm$0.14}
      & 80.41 {\scriptsize$\pm$0.14}
      & 63.97 {\scriptsize$\pm$0.46}
      & 65.35 {\scriptsize$\pm$0.42}
      & 31.68 {\scriptsize$\pm$1.33} \\[4pt]

    +P.-Gaussian
      & 96.32 {\scriptsize$\pm$0.02}
      & 90.96 {\scriptsize$\pm$0.17}
      & 79.56 {\scriptsize$\pm$0.37}
      & 68.79 {\scriptsize$\pm$0.36}
      & 65.56 {\scriptsize$\pm$0.39}
      & 32.21 {\scriptsize$\pm$0.93} \\

    +P.-Gaussian (Soft)
      & 96.44 {\scriptsize$\pm$0.14}
      & \textbf{91.36} {\scriptsize$\pm$0.21}
      & 79.86 {\scriptsize$\pm$0.14}
      & \textbf{68.97} {\scriptsize$\pm$0.27}
      & 65.43 {\scriptsize$\pm$0.53}
      & 33.53 {\scriptsize$\pm$2.11} \\[2pt]

    +Gaussian
      & 95.52 {\scriptsize$\pm$0.08}
      & 90.97 {\scriptsize$\pm$0.16}
      & 77.30 {\scriptsize$\pm$0.33}
      & 68.09 {\scriptsize$\pm$0.19}
      & 64.85 {\scriptsize$\pm$0.34}
      & \textbf{34.91} {\scriptsize$\pm$1.90} \\

    +Gaussian (Soft)
      & 95.65 {\scriptsize$\pm$0.14}
      & 91.09 {\scriptsize$\pm$0.08}
      & 77.24 {\scriptsize$\pm$0.10}
      & 68.16 {\scriptsize$\pm$0.23}
      & 64.93 {\scriptsize$\pm$0.50}
      & 34.35 {\scriptsize$\pm$1.24} \\[2pt]

    +RE
      & 96.56 {\scriptsize$\pm$0.19}
      & 87.48 {\scriptsize$\pm$0.43}
      & 80.68 {\scriptsize$\pm$0.14}
      & 64.39 {\scriptsize$\pm$0.61}
      & 66.25 {\scriptsize$\pm$0.47}
      & 31.40 {\scriptsize$\pm$1.70} \\

    +RE (Soft)
      & 96.70 {\scriptsize$\pm$0.21}
      & 87.59 {\scriptsize$\pm$0.41}
      & \textbf{80.74} {\scriptsize$\pm$0.20}
      & 64.58 {\scriptsize$\pm$0.60}
      & 66.05 {\scriptsize$\pm$0.40}
      & 31.67 {\scriptsize$\pm$1.78} \\

    +Soft RE (w)
      & 96.71 {\scriptsize$\pm$0.11}
      & 87.69 {\scriptsize$\pm$0.57}
      & 80.45 {\scriptsize$\pm$0.26}
      & 63.62 {\scriptsize$\pm$0.37}
      & 66.22 {\scriptsize$\pm$0.36}
      & 31.68 {\scriptsize$\pm$1.67} \\[2pt]

    +RC
      & 96.64 {\scriptsize$\pm$0.18}
      & 87.21 {\scriptsize$\pm$0.42}
      & 80.20 {\scriptsize$\pm$0.25}
      & 62.55 {\scriptsize$\pm$0.64}
      & 65.44 {\scriptsize$\pm$0.28}
      & 31.02 {\scriptsize$\pm$0.94} \\

    +RC (Soft)
      & \textbf{96.76} {\scriptsize$\pm$0.11}
      & 86.04 {\scriptsize$\pm$0.62}
      & 80.50 {\scriptsize$\pm$0.28}
      & 61.89 {\scriptsize$\pm$0.92}
      & \textbf{66.54} {\scriptsize$\pm$0.44}
      & 31.56 {\scriptsize$\pm$1.02} \\

    +RC (Soft) (w)
      & 96.63 {\scriptsize$\pm$0.19}
      & 85.54 {\scriptsize$\pm$0.65}
      & 80.65 {\scriptsize$\pm$0.21}
      & 61.52 {\scriptsize$\pm$0.94}
      & 66.39 {\scriptsize$\pm$0.32}
      & 30.86 {\scriptsize$\pm$1.31} \\

    \bottomrule
  \end{tabular}}
\end{table}

\setlength{\tabcolsep}{2.5pt}

\begin{table}[!t]
  \centering
  \caption{Ablation study applying various (soft) augmentations incrementally on top of standard, not softened TrivialAugment. All entries except TA show mean $\Delta$ $\pm$ standard deviation of $\Delta$ (relative to TrivialAugment). When multiple soft augmentations are applied, the reduced confidence is multiplied, which combines the adaptive label smoothing of all soft augmentations. P.-Gaussian means Patch Gaussian. Overall, the best augmentation combination varies from dataset to dataset and does not always involve adaptive label smoothing at all.}
  \label{tab:ablation}
  \resizebox{\textwidth}{!}{%
  \begin{tabular}{@{}l  rr  rr  rr@{}}
    \toprule
    & \multicolumn{2}{c}{CIFAR-10}
    & \multicolumn{2}{c}{CIFAR-100}
    & \multicolumn{2}{c}{TinyImageNet} \\
    \cmidrule(lr){2-3}\cmidrule(lr){4-5}\cmidrule(lr){6-7}
    Method 
      & Acc.      & Rob.
      & Acc.      & Rob.
      & Acc.      & Rob. \\
    \midrule
    TrivialAugment
      & 96.42 {\scriptsize$\pm$0.07}
      & 86.78 {\scriptsize$\pm$0.14}
      & 80.41 {\scriptsize$\pm$0.14}
      & 63.97 {\scriptsize$\pm$0.46}
      & 65.35 {\scriptsize$\pm$0.42}
      & 31.68 {\scriptsize$\pm$1.33} \\[4pt]

    +RC
      & +0.23 {\scriptsize$\pm$0.21}
      & +0.42 {\scriptsize$\pm$0.39}
      & -0.21 {\scriptsize$\pm$0.24}
      & -1.41 {\scriptsize$\pm$0.79}
      & +0.09 {\scriptsize$\pm$0.49}
      & -0.66 {\scriptsize$\pm$1.97} \\

    +RC (Soft)
      & +0.35 {\scriptsize$\pm$0.13}
      & -0.74 {\scriptsize$\pm$0.60}
      & +0.08 {\scriptsize$\pm$0.35}
      & -2.08 {\scriptsize$\pm$0.53}
      & +1.19 {\scriptsize$\pm$0.12}
      & -0.12 {\scriptsize$\pm$2.01} \\[3pt]

    +RC+RE
      & \textbf{+0.44} {\scriptsize$\pm$0.08}
      & +0.88 {\scriptsize$\pm$0.44}
      & +0.22 {\scriptsize$\pm$0.16}
      & -0.34 {\scriptsize$\pm$0.50}
      & +0.85 {\scriptsize$\pm$0.41}
      & +0.66 {\scriptsize$\pm$2.29} \\

    +RC+RE (Soft)
      & +0.33 {\scriptsize$\pm$0.10}
      & -0.14 {\scriptsize$\pm$0.97}
      & \textbf{+0.42} {\scriptsize$\pm$0.35}
      & -1.09 {\scriptsize$\pm$0.69}
      & \textbf{+1.54} {\scriptsize$\pm$0.40}
      & +0.59 {\scriptsize$\pm$1.65} \\[3pt]

    +RC+RE+P.-Gaussian
      & +0.15 {\scriptsize$\pm$0.09}
      & +4.18 {\scriptsize$\pm$0.33}
      & -0.51 {\scriptsize$\pm$0.17}
      & +4.49 {\scriptsize$\pm$0.46}
      & +0.83 {\scriptsize$\pm$0.49}
      & +0.10 {\scriptsize$\pm$1.94} \\

    +RC+RE+P.-Gaussian (Soft)
      & -0.31 {\scriptsize$\pm$0.14}
      & +3.73 {\scriptsize$\pm$0.17}
      & -1.12 {\scriptsize$\pm$0.23}
      & +4.30 {\scriptsize$\pm$0.46}
      & -0.14 {\scriptsize$\pm$0.44}
      & +1.78 {\scriptsize$\pm$2.45} \\[3pt]

    +RC+RE+Gaussian
      & -0.43 {\scriptsize$\pm$0.16}
      & \textbf{+4.39} {\scriptsize$\pm$0.22}
      & -2.43 {\scriptsize$\pm$0.19}
      & \textbf{+4.86} {\scriptsize$\pm$0.51}
      & +0.93 {\scriptsize$\pm$0.54}
      & \textbf{+3.47} {\scriptsize$\pm$1.78} \\

    +RC+RE+Gaussian (Soft)
      & -0.44 {\scriptsize$\pm$0.13}
      & +4.09 {\scriptsize$\pm$0.22}
      & -1.98 {\scriptsize$\pm$0.17}
      & +4.50 {\scriptsize$\pm$0.34}
      & +1.15 {\scriptsize$\pm$0.58}
      & +2.13 {\scriptsize$\pm$2.07} \\

    \bottomrule
  \end{tabular}}
\end{table}

\section{Discussion}
Our results confirm that adaptive label smoothing tied to a single-transform augmentation other than RC can enable more aggressive parameter settings and improve training with that augmentation. While we could show these benefits for RE, it was less significant for noise injections. In many cases, extensive search is needed to find the optimal parametrization that yields an advantage for adaptive label smoothing. Overall, our results suggest that practitioners that predominantly work with a single type of augmentation, like RE for occlusion-sensitive applications, should consider adaptive label smoothing. 

By contrast, when multiple transformations are applied together (e.g. in TrivialAugment or mixed schemes), the diversity of distortions alone suffices to regularize the model, and adaptive smoothing yields only marginal or no additional benefit. In fact, overly aggressive smoothing generally appears to reduce robustness to corruption, as is also true of the leading soft RC method. A reason could be that excessive label uncertainty on strong transformations hinders learning invariance to these transformations.

We note several simplifying assumptions in our current framework. Even though the label smoothing is adaptive, we treat every transformation type and magnitude as inducing a uniform drop in label confidence, irrespective of the specific image, class, or dataset context. Future work could refine this by conditioning smoothing factors on image content or class, or by integrating auxiliary label confidence estimators.

\section{Conclusion}
This study proposed multiple approaches to extend adaptive label smoothing beyond random cropping to work with other aggressive, state-of-the-art augmentations. It leveraged human‐vision data and proxy‐model accuracy to map transformation magnitude to adaptive label confidence. This approach enhances the efficacy of data augmentation schemes building on individual transformations such as Random Erasing, allowing a more aggressive parametrization without miscalibrating the model. However, the advantage vanishes when a heterogeneous set of transformations is deployed on the image data. Our findings narrow down the regime in which adaptive label smoothing is valuable: namely, when a singular transformation dominates the augmentation distribution.

%
%
%
%
\bibliographystyle{splncs04}
\bibliography{egbib}

\newpage
\appendix
\appendix
\section{Appendix}
\subsection{Training parameters}\label{sec:training_setup}

Table \ref{tab:training_params} summarizes the parameters of our training process for the classification models that are fixed for all experiments.

\begin{table}[htb]
  \centering
  \caption{Training parameters}
  \label{tab:training_params}
  \begin{tabular}{@{}ll@{}}
    \toprule
    Parameter                          & Value                                \\ 
    \midrule
    Learning rate schedule             & Cosine Annealing                 \\
    Initial learning rate              & 0.1                                   \\
    Epochs                             & 300                                   \\
    Batch size                         & 128                                   \\
    Optimizer                          & SGD                                   \\
    Momentum                           & 0.9                                   \\
    Weight decay                       & $10^{-4}$                     \\
    Dropout rate                       & 0.3                                   \\
    Data Normalization                 & None                     \\
    Random Erasing value               & Random standard gaussian                 \\ 
    Random Erasing aspect value        & uniformly drawn from [0.3, 3.3] \\
    Data Augmentations                 & Random horizontal flips            \\ 
    \bottomrule
  \end{tabular}
\end{table}

\end{document}